\documentclass[9pt]{article}
\pdfoutput=1
\usepackage{spconf}
\usepackage{pgfplotstable}

\usepackage[english]{babel}
\usepackage[utf8x]{inputenc}
\usepackage[T1]{fontenc}

\usepackage{amsmath,amssymb,mathrsfs,bbm}
\usepackage{graphicx}
\usepackage[colorinlistoftodos]{todonotes}
\usepackage{multirow}
\usepackage{makecell}
\usepackage[lined,linesnumbered,ruled]{algorithm2e}

\usepackage{psfrag,epsfig,graphics}
\usepackage{amsmath,amsthm,amssymb,multirow}
\usepackage{amssymb}   
\usepackage{url}  
\usepackage{xstring}
\usepackage[noadjust]{cite}

\usepackage{mathabx} 

\DeclareSymbolFontAlphabet{\amsmathbb}{AMSb}%

\newcommand{\cp}[1]{\ifmmode {\mathcal{#1}}\else ${\mathcal{#1}}$\fi}
	
\newcommand{\bA}{\boldsymbol{A}}

\newcommand{\bC}{\boldsymbol{C}}

\newcommand{\bX}{\boldsymbol{X}}

\newcommand{\bb}{\boldsymbol{b}}

\newcommand{\bs}{\boldsymbol{s}}
\newcommand{\bu}{\boldsymbol{u}}
\newcommand{\bv}{\boldsymbol{v}}
\newcommand{\bx}{\boldsymbol{x}}

\newcommand{\bbR}{\mathbb{R}}

\newcommand{\btheta}{\boldsymbol{\theta}}

\newcommand{\bTheta}{\boldsymbol{\Theta}}

\usepackage{color}  

\definecolor{darkgreen}{rgb}{0., 0.4, 0.}

\usepackage{booktabs}
\usepackage{comment}
\def\BibTeX{{\rm B\kern-.05em{\sc i\kern-.025em b}\kern-.08em
    T\kern-.1667em\lower.7ex\hbox{E}\kern-.125emX}}

\usepackage{tikz}
\usetikzlibrary{bayesnet}
\usetikzlibrary{arrows}
\usetikzlibrary{positioning}

\title{Inv-SENnet: Invariant Self Expression Network for clustering under biased data\\
}

\name{Ashutosh Singh$^{\dagger,*}$, Ashish Singh$^{\ddagger,*}$, Aria Masoomi$^\dagger$, Tales Imbiriba$^\dagger$, Erik Learned-Miller$^\ddagger$,  Deniz Erdo{\u{g}}mu{\c{s}}$^\dagger$
\address{
$^\dagger$ Dept. of Electrical \& Computer Engineering, Northeastern University, Boston, MA, USA\\
$^\ddagger$ College of Information \& Computer Sciences, University of Massachusetts Amherst, MA, USA} 
 \thanks{This work was supported by NSF (1947972).}
\thanks{$^*$Indicates shared first authorship. Correspondence should be addressed to Ashutosh Singh (singh.ashu@northeastern.edu).}
} 

\begin{document}
\ninept
\maketitle
\begin{abstract}

Subspace clustering algorithms are used for understanding the cluster structure that explains the dataset well. These methods are extensively used for data-exploration tasks in various areas of Natural Sciences. However, most of these methods fail to handle unwanted biases in datasets. For datasets where a data sample represents multiple attributes, naively applying any clustering approach can result in undesired output. To this end, we propose a novel framework for jointly removing unwanted attributes (biases) while learning to cluster data points in individual subspaces. Assuming we have information about the bias, we regularize the clustering method by adversarially learning to minimize the mutual information between the data and the unwanted attributes. Our experimental re-
sult on synthetic and real-world datasets
demonstrate the effectiveness of our approach.

\end{abstract}

%

\section{Introduction}

   Most real world datasets often carry information arising from several attributes. Some of these attributes are of no importance and do not represent the goal of the task. Ideally, these attributes should be uncorrelated to the features of the dataset that are important to fullfil the task. But due to various reasons in many real world datasets, these biases appear to be highly correlated with these features of importance. This mostly appears due to the nature of the data collection pipeline\cite{huang2008labeled}. When such datasets are used for training data driven models the results thus acquired also get affected by these biases. 
    
     
    This problem of performing inference when the data contains biases is studied under the broader umbrella of bias mitigation. Most of the past work in this area is aimed at solving this problem with supervision\cite{kim2019learning}\cite{alvi2018turning}\cite{attenberg2015beat} i.e. when there are labels pertaining to the desired task present during the training time. In many naturally appearing datasets, it is very hard to label the data particularly based on the factors of importance. Therefore to learn the patterns from such datasets unsupervised methods, \textit{e.g.} clustering, are used. But when biases are present in the data the result of such unsupervised methods is often highly correlated with the biases. In this paper, we try to address this problem. 
    
    Among the data-driven approaches for clustering, subspace clustering shows great promise. In subspace clustering the assumption is that the high dimensional data lie on a union of low dimensional subspaces. The objective here is to find separate subspaces for separate clusters of data points. Among the family of subspace clustering algorithms Self-expression based algorithms form the state of the art\cite{zhang2021learning}. Self expression imposes the constraint that every data point in the dataset can be explained through a linear combination of all the other data point in the dataset,
    \begin{equation}
        x _i= \sum_{i \neq j} c_{i,j}x_j
    \end{equation}
    Here $c_{ij}$ represents the coefficient of the $j^{th}$ datapoint w.r.t. its contribution to reconstruct the $i^{th}$ datapoint. One of the reason why self expression based methods are popular is because of the subsapce preserving property of the coefficient matrix achieved under certain regularisation function \cite{elhamifar2013sparse}\cite{favaro2011closed}\cite{soltanolkotabi2012geometric}\cite{soltanolkotabi2014robust}\cite{you2020self}. This means that $c_{ij}$ is only non-zero when the $i^{th}$ and $j^{th}$ datapoints are in the same subspace. Most recent advances in subspace clustering literature have focused on scalability and out of sample clustering using neural networks. But even these algorithms fail under the presence of biases.

 More often we have labels pertaining to these sources of biases. Under the assumption that the labels of the biases are known during the training, we propose an information-theoretic inspired way of jointly learning the clustering while ignoring the bias confounds in the data. We evaluate our proposed method over synthetic and naturally occurring image datasets and show superior performance than the current state of the art.

The remaining paper is organized as follows. Section 2 presents the problem statement, section 3 presents the proposed solution, section 4 presents the experiments and section 5 presents the discussion.



\section{Problem Statement}
Most clustering algorithms are required to be trained on the whole dataset to find clusters. As the size of the dataset increases, they become hard to scale. Scalable Subspace clustering algorithms show great promise in solving this problem \cite{elhamifar2012see}\cite{you2016scalable}\cite{peng2013scalable}. The method proposed in \cite{zhang2021learning}, based on self-expression, scales well for large datasets and shows great performance on out-of-sample data points. But under the presence of bias, these methods fail on out-of-sample data points resulting in learning wrong self-expression coefficients. In this paper, we focus on this problem and look toward improving the existing method. Below we formulate the problem of biases in clustering.

\subsection{Formulation} \label{sec:probForm}
Let $\bx \in \mathbb{R}^d$ be the input data and $\bX = [\bx_1, \ldots, \bx_n] \in \mathbb{R}^{d\times n}$ represent the input data matrix where the $x_j$ represents different data points while the rows represent the dimensions. We make the assumption that the data, \emph{i.e.}, the columns in $\bX$ lie on the union of $k$ low-dimensional subspaces $\bx_j \in \cup_{i=1}^{k}\mathcal{X}_i$. 
Let $\bb = [b_1,\ldots, b_2]$ represent the set of bias labels, where $b_j\in \{0,1\}$ indicate the existence of a bias factor for every $\bx_j$. Based on the self-expressive models \cite{elhamifar2013sparse} we can define each data point in $\bX$ as a linear combination of all the other data points as \cite{elhamifar2013sparse}, 
\begin{equation}
    \bx_j = \sum_{i\neq j} c_{ij}\bx_i
\end{equation}
Here $c_{ij}$ represents the elements of the matrix $\bC$, self-expressive coefficient matrix, satisfying the following,
\begin{equation}
    \bX = \bC \bX,
\end{equation}
where the diagonal element of $\bC$ equals 0.
Hence learning of the self expressive coefficient matrix could be formulated as the following optimisation problem,
\begin{equation}\label{eq:Loss1}
\min_{\bC} \mathcal{L}(\bX, \bC \bX) - \mathcal{R}(\bC) \,,
\end{equation}
where $\mathcal{L}$ and $\mathcal{R}$ represent cost and regularization terms, respectively.
Therefore the $\bC$ matrix, subspace preserving, can be further used to get the affinity matrix $\bA = |\bC| + |\bC^\top|$. We can obtain the clustering labels $\hat{\bs}$ over $\bX$ using the affinity matrix in the spectral clustering algorithm \cite{ng2001spectral}.

Now, let $\bs$ be the set of true (desired) cluster labels, and $\mathcal{I}(\bx,\bx^\prime)$ be the mutual information between $\bx$ and $\bx^\prime$. Suppose $\hat{\bs}$ are the cluster labels based on the $\bC$ matrix we get from solving \eqref{eq:Loss1}. Then due to the presence of biases present in the data we can very possibly observe,
\begin{equation}\label{eq:mi_ineq}
    \mathcal{I}(\hat{\bs}, \bb) \gg \mathcal{I}(\bs, \bb)
\end{equation}
The effect of bias on classification task is already well discussed and studied in literature \cite{kim2019learning}. Therefore in task where supervision is not present, its very likely that presence of strong biases in the data would lead to clustering algorithm output getting biased too as formulated in~\eqref{eq:mi_ineq}. This would likely result in $\hat{\bs} \neq \bs$.
In this contribution we aim at solving the problem in~\eqref{eq:Loss1} while enforcing bias mitigation as expressed in Eq.~\eqref{eq:mi_ineq}.
To solve this problem we propose an information theoretic based approach in the next section.

\section{Bias mitigation for Self-expressive networks}

In this section we discuss the elements of our proposed solution and finally discuss the architecture and optimisation for Inv-SENnet.

\begin{figure}[t]
    \begin{center}
    \includegraphics[width=0.8\columnwidth]{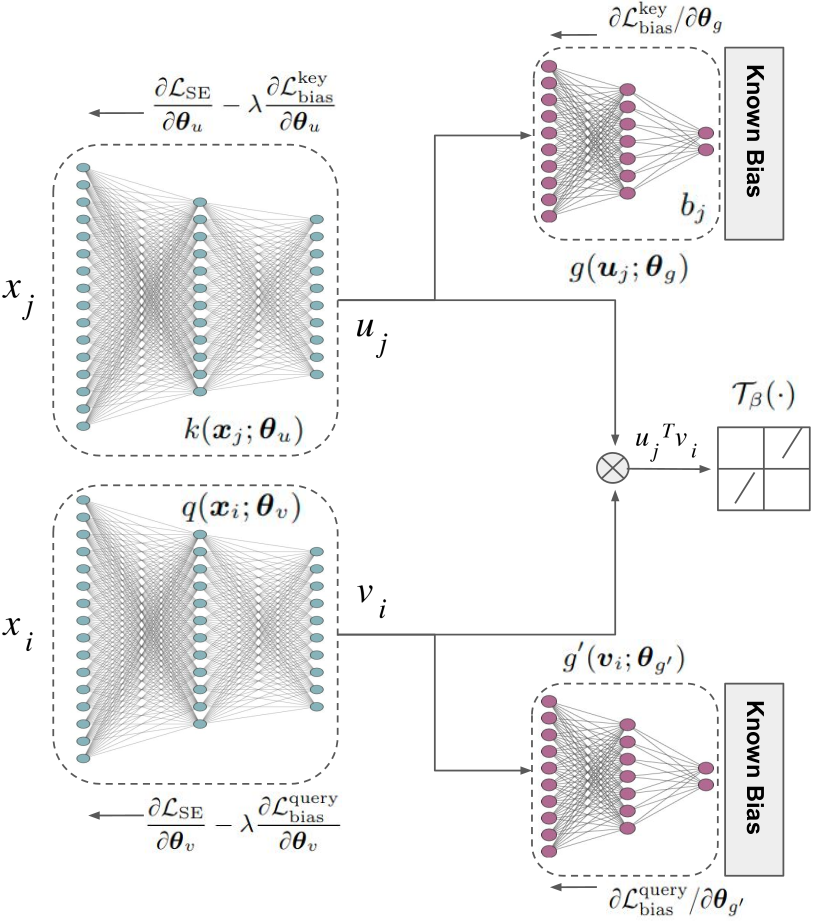}
    \end{center}
    \vspace{-0.3cm}
    \caption{Architecture for INV-SENnet. Here $k(\bx;\btheta_u)$ and $q(\bx;\btheta_v)$ be key and the query net respectively. $g(\bu_j;\btheta)$ and $g^{\prime}(\bv_i;\btheta_{g^{\prime}})$ are the bias classifiers. }
    \label{fig:rho}
\end{figure}

\subsection{Self Expressive Network}
In literature there are many optimisation framework already proposed that try to solve \eqref{eq:Loss1} to learn the self expression coefficient matrix $C$. Most of the past works have tried to solve \eqref{eq:Loss1} optimising  the following loss, 
\begin{equation}
   \mathcal{L}_\mathrm{SE} = \min_{c_{ij}, i\neq j} \|\bx_j - \sum_{i \neq j} c_{ij}\bx_i\|^2_{2} -\sum_{i \neq j}r( c_{ij})
   \label{eq:se_loss}
\end{equation}

where $r(\cdot)$ represents the regularisation term based on elastic net regularisation \cite{you2016oracle}. The role of $r(\cdot)$ is to provide denser subspace preserving coefficient matrix. 
Let $f(\cdot) : \mathbb{R}^D \times \mathbb{R}^D \rightarrow \mathbb{R}$ be the function, parameterised by $\bTheta$ taking arguments $\bx_i$ and $\bx_j$, and giving self-expression coefficients $c_{i,j}$ in matrix $\bC$, \textit{i.e.} $f(\bx_i,\bx_j;\bTheta) = c_{ij}$. Therefore we can rewrite \eqref{eq:se_loss} as the following optimisation,

\begin{equation} \label{eq:cost_SE}
    \mathcal{L}_\mathrm{SE} = \min_{\bTheta}\frac{\gamma}{2}\|\bx_j - \sum_{i \neq j} f(\bx_i, \bx_j;\bTheta)\bx_i\|^2_{2} - r(f(\bx_i,\bx_j;\bTheta))
\end{equation}

In \cite{zhang2021learning}, the authors propose the SENnet, which models $f(\bx_i, \bx_j; \bTheta)$ as neural network expressed as
\begin{align*}
    f(\bx_i,\bx_j; \bTheta) = \alpha\mathcal{T}_{\beta}(\bu^\top \bv)\\
    \bu_i = k(\bx_i;\btheta_u) \in \mathbb{R}^p\\
    \bv_i = q(\bx_j;\btheta_v) \in \mathbb{R}^p
\end{align*}
where,
\begin{equation}
\mathcal{T}_{\beta}(\cdot) = \text{sign}(t)\max(0, |t| - \beta) .
\end{equation}
Here $k$ and $q$ represents the key and the query networks, while $\mathcal{T}$ is a soft threshholding operator. $\bTheta = \{\btheta_u, \btheta_v, \beta\}$ represents the learnable parameters of the model. 
We highlight that SENet is not designed to cluster the data when bias is present in the data. 
We improve upon their model by adding a bias mitigation mechanism. The resulting methodology leads to a learning algorithm capable of performing clustering and mitigation jointly as described next.

\subsection{Bias Mitigation}
As discussed in Section~\ref{sec:probForm}, the presence of biases in the training dataset leads to biases-dependent solutions for the optimization problem, and, therefore, leads to biased clustering. Below we present the proposed bias mitigation strategy.
%
%
For this, we assume that the training dataset contains labels for the bias classes. Note, however, that biases labels are not required for the test set. Therefore the goal here is to propose a way of learning $\bTheta$ s.t. $c_{ij}$ is invariant to the presence of bias.

Referring to the proposed architecture in  Fig.~\ref{fig:rho} we define $g(k(\bx;\btheta_u); \btheta_g): \bbR^p \to \bbR$ and $g'(q(\bx;\btheta_v); \btheta_g'): \bbR^p \to \bbR$ as the bias classification functions over $\bu$ and $\bv$ that gives the $\bb$ labels for the corresponding $\bx$. 
Therefore the optimisation framework in~\eqref{eq:cost_SE} can be modified as:
\begin{equation}
    \mathcal{L} = \mathcal{L}_{\mathrm{SE}} - \lambda (\mathcal{I}(\bu, \bb) + \mathcal{I}(\bv, \bb)) .
    \label{eq:all_loss}
\end{equation}

Here $\lambda\in\mathbb{R}_+$ is a scalar hyper-parameter controlling the bias mitigation component in the loss. Estimating $\mathcal{I}(\cdot)$ requires the joint distributions $p(\bb,\bv)$ and $p(\bb,\bu)$. This makes the problem hard to solve and often cases intractable. Mutual information can also be written in terms of the conditional entropy such that 
\begin{align*}
\mathcal{I}(\bb,\bv) = \mathcal{H}(\bb) - \mathcal{H}(\bb|\bv), \qquad \mathcal{I}(\bb,\bu) = \mathcal{H}(\bb) - \mathcal{H}(\bb|\bu) .
\end{align*}
Here, $\mathcal{H}$ represents the entropy i.e. $\mathcal{H}(\bb) = log(P(\bb))$.
Note that $\mathcal{H}(\bb)$ can be ignored when optimizing~\eqref{eq:all_loss} since the bias $\bb$ is independent of the model parameters $\bTheta$. 
To compute conditional entropies conditional densities such as $P(\bb|\bu)$ and $P(\bb|\bv)$ are required, but can be difficult to obtain.
In~\cite{kim2019learning} the authors approximated the conditional distribution using a variational distribution $Q$.
Following a similar reasoning we can define two distributions $Q$ and $Q^\prime$ to approximate $P(\bb|\bu)$ and $P(\bb|\bv)$, respectively. Now, let $\btheta_g$ and $\btheta_{g^\prime}$ be the parameters of $Q$ and $Q^\prime$. Thus, each parameter can be learned by minimizing the Kullback–Leibler ($\mathcal{D}_{KL}$) divergence between the respective distributions. For $\btheta_g$ we have:
\begin{equation}
    \min_{\btheta_g}\mathcal{D}_{KL}[Q(g(k(\bx;\btheta_u);\btheta_g)|k(\bx;\btheta_u)) \| P(\bb|k(\bx;\btheta_u))]
    \label{eq:KL divergence}
\end{equation}
while an equivalent problem can be solved for $\bv = q(\bx;\btheta_v)$, omitted here for sake of space. 

Now, we can define the bias classification optimization problem by combining a cross-entropy loss and the KL in~\eqref{eq:KL divergence} as a regularization term. The goal of this optimization is to not only make the bias classifier able to classify the bias class accurately, but also to make the $k(\bx;\btheta_u)$ invariant of the bias. To solve the second objective we can impose a regularisation on $\btheta_u$ aiming at maximizing the negative of \eqref{eq:KL divergence}, hence making it hard for the $g(\cdot)$ to classify the bias. Therefore the optimisation becomes a $\min\max$ problem,
\begin{align}
    \mathcal{L}_\mathrm{bias}^\mathrm{key} &= \min_{\btheta_u}\max_{\btheta_g}\mathbb{E}_{\bx \sim P_X(\cdot)}[\mathbb{E}_{\tilde{\bb} \sim Q(\cdot|k(\bx;\btheta_u))}[logQ(\tilde{\bb}|k(\bx;\btheta_u))]] \nonumber\\ 
    & - \mu \mathbb{E}_{\bx \sim P_X(\cdot)}[\mathcal{L}_c(\bb, g(k(\bx;\btheta_u);\btheta_g)]
    \label{eq:bias}
\end{align}
where $\mathcal{L}_c(\cdot)$ represents the cross entropy loss. The second term in \eqref{eq:bias} is the relaxation term for the KL divergence condition in \eqref{eq:KL divergence}. Similar discussion is presented in \cite{kim2019learning}. 
Equivalently, we can define the minimax problem $\mathcal{L}_\mathrm{bias}^\mathrm{query}$ for optimizing $\btheta_v$ and $\btheta_{g^\prime}$, also omitted in this manuscript.



Let $\bTheta_g = \{\btheta_g, \btheta_{g^{\prime}}\}$ represent the set of all parameters of the bias classifiers. Therefore, combining (\ref{eq:bias}), for both key and query networks, and (\ref{eq:se_loss}) we can rewrite~\eqref{eq:all_loss} as:

\begin{align}
    \mathcal{L} &= \mathcal{L}_{\textrm{SE}} - \lambda(\mathcal{L}_{\textrm{bias}}^{\textrm{key}} + \mathcal{L}_\textrm{bias}^\textrm{query}) \nonumber\\
    &= \min_{\bTheta}\max_{\bTheta_g}\frac{\gamma}{2n}\sum_{j}[\|\bx_j - \sum_{i \neq j} \mathcal{T}_{\beta}(k(\bx_j;\btheta_u)^Tq(\bx_i;\btheta_v))\bx_i\|_{2}^{2} \nonumber\\ 
    &\quad - r(\mathcal{T}_{\beta}(k(\bx_j;\btheta_u)^Tq(\bx_i;\btheta_v)))] \nonumber\\
    &\quad - \lambda\mathbb{E}_{\bx \sim P_{\bX}(\cdot)}[\mathbb{E}_{\tilde{\bb} \sim Q(\cdot|k(\bx;\btheta_u))}[\log Q(\tilde{\bb}|k(\bx;\btheta_u))]] \label{eq:loss_final}\\ 
    &\quad + \mu \mathbb{E}_{\bx \sim P_{\bX(\cdot)}}[\mathcal{L}_c(\bb, g(k(\bx;\btheta_u);\btheta_g)] \nonumber\\
    &\quad - \lambda\mathbb{E}_{\bx \sim P_{\bX}(\cdot)}[\mathbb{E}_{\tilde{\bb} \sim Q'(\cdot|q(\bx;\btheta_v))}[\log Q'(\tilde{\bb}|q(\bx;\btheta_v))]] \nonumber\\ 
    &\quad + \mu \mathbb{E}_{\bx \sim P_{\bX(\cdot)}}[\mathcal{L}_c(\bb, g^{\prime}(q(\bx;\btheta_v);\btheta_g^{\prime})] .\nonumber
\end{align}


\subsection{Training}

In \eqref{eq:loss_final} we present the overall loss function for the architecture proposed in Fig. \ref{fig:rho}. In practice, we use two optimizers; one to solve the inner maximisation and the second one to solve the outer minimisation. First we compute the forward pass of the data through $k(\bx_j;\btheta_u)$ and $q(\bx_i;\btheta_v)$ and compute $\mathcal{L}_{\mathrm{SE}}$. At the same time we also pass both $\bu_j$ and $\bv_i$ through their respective bias classifiers $g(\cdot)$ and $g^{\mathrm{\prime}}(\cdot)$ to get the bias classification. We then compute the $\partial{\mathcal{L}_{\mathrm{bias}}^{\mathrm{key}}}/\partial{\btheta_g}$ and $\partial{\mathcal{L}_{\mathrm{bias}}^{\mathrm{query}}}/\partial{\btheta_{g^{\prime}}}$ and backpropagate through the bias classifiers $g(\cdot)$ and $g^{\prime}(\cdot)$ respectively. Using the gradient reversal technique \cite{ganin2016domain} together with the gradient of $\mathcal{L}_{\mathrm{SE}}$ we compute,
\begin{align*}
\nabla\btheta_u = \frac{\partial{\mathcal{L}_{\mathrm{SE}}}}{\partial{\btheta_u}} -\lambda\frac{\partial{\mathcal{L}_{\mathrm{bias}}^{\mathrm{key}}}}{\partial{\btheta_u}}, \qquad \nabla\btheta_v = \frac{\partial{\mathcal{L}_{\mathrm{SE}}}}{\partial{\btheta_v}} -\lambda\frac{\partial{\mathcal{L}_{\mathrm{bias}}^{\mathrm{query}}}}{\partial{\btheta_v}} .
\end{align*}
for key and query network respectively. At the beginning of the training, the bias classifiers converges quickly. But as the training goes on the $C$ matrix becomes invariant of the biases and the bias classifier starts performing poorly. This happens because the $\mathcal{I}(\bu,\bb)$ and $\mathcal{I}(\bv,\bb)$ goes down \textit{i.e} the key and the query net becomes good at unlearning the biases. 

\begin{figure*}[tbh]
    \centering
    \includegraphics[width=0.9\linewidth]{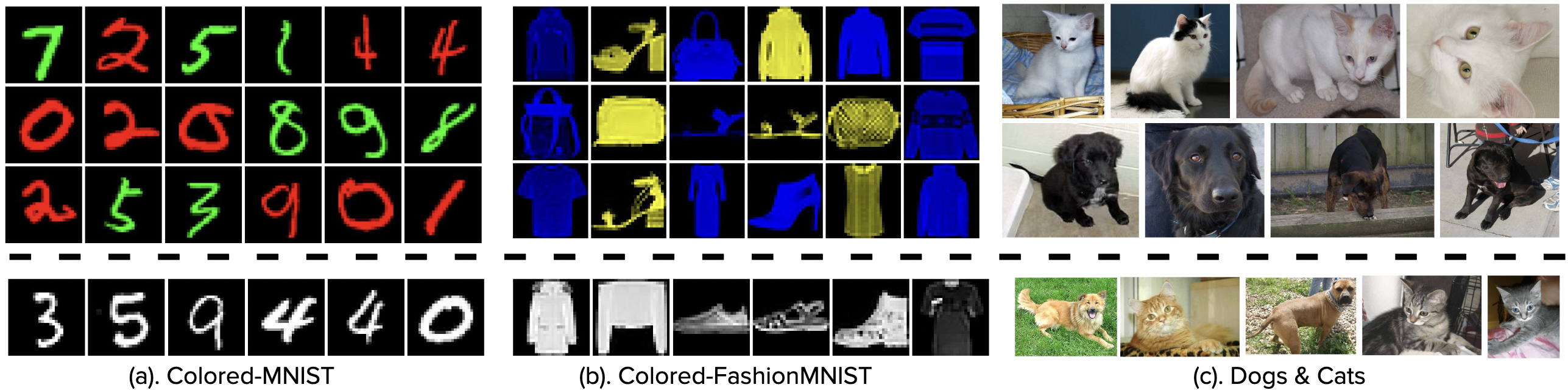}
    \vspace{-0.3cm}
    \caption{Examples of Datasets with bias [Top-Rows: Biased set; Bottom-Rows: Unbiased set]. We modified the MNIST and FashionMNIST dataset by randomly adding color ([Red ,Green] for MNIST and [Blue, Yellow]  for FashionMNIST) to each dataset. For Dogs and Cats dataset, we sub-sampled dark dog and bright cat images from the original dataset.}
    \label{fig:dataset}
\end{figure*}

\section{Experiment}
We conduct experiments on real-world benchmarks to evaluate our model on clustering biased data. We evaluate our method first on image dataset and then discuss a case study on an fMRI dataset.
\subsection{Setup}
\textbf{Network Architecture:} 
We formulate query, key and bias networks as an MLP. For query and key network, we create a three-layer MLP with ReLU and tanh(·) as the activation functions. The number of hidden units in each layer of these MLP are $\{1024, 1024, 1024\}$, and the output dimension is 1024. For the Bias networks, we use a three-layer MLP of dimensions $\{1024, 512, 256\}$ followed by the classification layer. We use ReLU activation function and batch normalization layer between each fully-connected layer. We apply softmax function to the output of the classification layer to compute class probability scores. To optimize our model, we use Adam \cite{kingma2014adam} optimizer with a constant learning rate of $1e−3$ for the query and key networks and $1e-4$ for the bias networks.
\\
\\
\textbf{Datasets:}
To evaluate our method, we consider two scenarios where our data samples are biased. In the first setup we intentionally add bias information to existing benchmarks (MNIST \cite{lecun1998gradient} and FashionMNIST \cite{xiao2017fashion}). In the second case, we consider a natural setting (Dogs and Cats dataset) where the bias information is inherently present in the data in the form of undesired attribute. Figure\ref{fig:dataset} shows some of the samples from our constructed datasets.\\
\textbf{Setting 1:} We select MNIST and FashionMNIST dataset for evaluation. Both benchmarks consists of grayscale image. We color each image red or green for MNIST and blue and yellow for FashionMNIST to introduce spurious correlations. We color each image in a manner that strongly correlates with original class labels \cite{arjovsky2019invariant}. This allows us to effectively measure the generalization properties of a given method. 
We create our training set in the following manner:
We first assign a binary label $y$ to the image based on the digit/class i.e. we set $y = 0$ for digits $0-4$ (classes $0-4$ for FashionMNIST) and $y = 1$ for digits $5-9$ (classes $5-9$ for FashionMNIST ). We then flip the label with $25\%$ probability. After that we color the image as one of the color according to its (possibly flipped) label. We flip the color with a probability $e$. For both benchmarks we set $e = 40\%$.
. We use scattering convolution transform\cite{bruna2013invariant} to generate features from MNIST and fashionMNIST similar to \cite{zhang2021learning} for optimal comparison.
\\
\textbf{Setting 2:} For this setting, We select the Dogs and Cats dataset from the kaggle competition \cite{Kaggle} for evaluation. We followed the similar setting from \cite{kim2019learning}, wherein, we subsample the original training set consisting of 25K images to create two subsets: a biased subset with dark colored dog images and bright colored cat images, and another set without those type of images. In total, our biased  and unbiased subset contains 6378 and 8125 images respectively. For feature extraction we used \cite{yu2020learning} as it learns to represent features in a union of subspaces.
\\
\\
\textbf{Metrics:}
For quantitative evaluation, we consider clustering accuracy (ACC), normalized mutual information (NMI) \cite{strehl2002cluster} and adjusted rand index (ARI) \cite{hubert1985comparing}. These metrics are commonly used in the literature to evaluate clustering methods.

\subsection{Results}
We show our experimental results in tables \ref{tab:results1} and \ref{tab:results2}. We evaluate vanilla SENet and INV-SENnet model for two settings: (a). out-of-distribution clustering and (b). Mix-domain clustering. In both the tables $N_i$ represents the number of training samples. For the first experiment, we want to test the generalizability of the learned self expression coefficient matrices. To this end, we first train the model  on the biased version of each dataset and then test it on the unbiased version. If a method is optimized to directly learn self expressive coefficients using the data without handling the bias information, it should fail to generalize to out-of-distribution samples where the spurious correlation might be absent. From the results in table \ref{tab:results1}, we can observe that having spurious correlation in a dataset can negatively affect clustering performance. This is evident from the results of standard SENet model, which fails to accurately cluster MNIST and FashionMNIST data based on ground-truth categories. The model performance is only marginally better than random chance ($10\%$ clustering accuracy for MNIST and FashionMNIST; $50\%$ clustering accuracy for Dogs and Cats). In contrast, we see considerable improvement of our proposed method over the standard model. In particular, we see a significant increase in all the three metrics for MNIST, while for FashionMNIST we see moderate improvements.For Dogs and Cats dataset, we again see improvement in performance when incorporating invariance to the bias information. 
 In the second experiment, we train and test on the same set that contains both biased and unbiased data in equal proportions. We perform this experiment to evaluate clustering when a dataset both biased and unbiased samples. We see a significant drop in all three of the metric for all of the benchmarks when comparing to Table \ref{tab:results1}. A very possible reason for this could be the mix of domains during the test times. Even then we see that INV-SENnet preforms better than SENnet on all the metrics. We see a significant improvement in the NMI score while comparing both the models. This further signifies that the clustering represents more of the desired attribute and less of the bias.

\begin{table}[thb]
\vspace{-0.5cm}
\footnotesize
\caption{Out-of-distribution clustering performance}
\centering
\renewcommand{\arraystretch}{1}
\setlength{\tabcolsep}{3.3pt}
\begin{tabular}{llcccc}
\bottomrule
Dataset & Method & $N_i$ & ACC (\%) & NMI. (\%) & ARI (\%) \\
\toprule	
\bottomrule

\multirow{4}{*}{MNIST}&\multirow{2}{*}{SENnet}  & 10000 & 41.79 & 33.34 & 21.05 \\
&& 60000 & 36.32 & 25.14 & 16.14 \\ 
&\multirow{2}{*}{Inv-SENnet} & 10000 & 68.44 & 60.55 & 45.23 \\ 
&& 60000 & 78.04 & 67.53 & 58.56 \\
\midrule
\multirow{4}{*}{Fashion-MNIST}&\multirow{2}{*}{SENnet}  & 10000 & 38.47 & 30.15 & 19.51 \\
&& 60000 & 46.73 & 42.46 & 31.18 \\ 
&\multirow{2}{*}{Inv-SENnet} & 10000 & 50.73 & 42.15 & 30.14 \\ 
&& 60000 & 56.91 & 44.68 & 34.41 \\
\midrule
Dogs and Cats&
SENnet & 6738 & 64.07 & 07.59 & 06.94 \\ 

&Inv-SENnet & 6738 & 78.53 & 25.32 & 32.4
 \\ 

\toprule		
\end{tabular}
\label{tab:results1}
\vspace{-0.5cm}
\end{table}

\begin{table}[thb]
\footnotesize
\caption{Mix-domain clustering performance}
\centering
\renewcommand{\arraystretch}{1}
\setlength{\tabcolsep}{3.3pt}
\begin{tabular}{llcccc}
\bottomrule
Dataset & Method & $N_i$ & ACC (\%) & NMI. (\%) & ARI (\%) \\
\toprule	
\bottomrule

\multirow{2}{*}{MNIST} &SENnet  & 15000 & 12.75 & 02.66 & 00.13 \\ 
&Inv-SENnet & 15000 & 35.29 & 25.84 & 11.31 \\ 
\midrule
\multirow{2}{*}{Fashion-MNIST}&SENnet  & 15000 & 20.37 & 13.50 & 02.39 \\
&Inv-SENnet & 15000 & 33.70 & 30.49 & 14.75 \\ 
\midrule
\multirow{2}{*}{Dogs and Cats}&
SENnet & 14953 & 32.68 & 44.97 & 22.90 \\ 

&Inv-SENnet & 14953 & 40.43 & 52.99 & 26.27
 \\ 

\toprule		
\end{tabular}
\label{tab:results2}
\vspace{-0.5cm}
\end{table}

\section{Conclusion}
In this work we formulated a bias mitigation strategy aiming at minimizing the mutual-information between bias and features. We applied this strategy for subspace clustering when the data is confounded by identifiable biases. The proposed InvSENnet was effective in learning bias-invariant subspace clustering,
reducing the influence of biases in different simulated and real datasets, clearly outperforming its version without bias mitigation capability.
Future work will focus on extending this work to datasets with modalities having multiple confounds such as EEG\cite{wang2018unsupervised}\cite{akbar2022post}and fMRI\cite{singh2021variation}\cite{barrett2017theory}. 


\bibliographystyle{IEEEtran}
\bibliography{BiasInvarSSC.bib}

\end{document}